\documentclass{article}

\usepackage[preprint]{neurips_2023}

\usepackage[utf8]{inputenc} %
\usepackage[T1]{fontenc}    %
\usepackage{hyperref}       %
\usepackage{url}            %
\usepackage{booktabs}       %
\usepackage{amsfonts, amsmath, amssymb, mathrsfs, mathtools}       %
\usepackage{nicefrac}       %
\usepackage{microtype}      %
\usepackage{xcolor}         %
\usepackage{csquotes}       %
\usepackage{graphicx}
\usepackage{subcaption}
\usepackage[most]{tcolorbox}
\usepackage{tikzscale}
\usepackage{soul}
\usepackage{listings}
\usepackage{tikz}
\usepackage{svg}
\usepackage{epsfig}
\usepackage{xspace}
\usepackage{algorithm}
\usepackage{algorithmic}
\usepackage[font=footnotesize]{caption}
\usepackage{bbm}
\usepackage{lipsum}
\usepackage[scale=.75]{miama}
\tcbuselibrary{breakable, skins}
\usepackage{spverbatim}

\graphicspath{{shrunk_figures/}{figures/}}

\hypersetup{
    colorlinks=true,
    linkcolor=blue,
    citecolor=blue,
    urlcolor=blue,
}
\DeclareMathOperator{\dataset}{\mathcal{D}}

\DeclareMathOperator*{\argmin}{arg\,min}
\newcommand{\finallayer}{l_\text{final}}
\newcommand{\loss}{L}

\newcommand{\model}{M}
\newcommand{\mediator}{v}

\DeclareMathOperator{\vol}{V}
\newcommand{\EPC}{effective parameter count\xspace}
\newcommand{\epcsymbol}{N_\text{eff}}
\newcommand{\pcsymbol}{N}
\newcommand{\dofsymbol}{N_\text{free}}
\newcommand{\experimentpaper}{\cite{experimentpaper}\xspace}

\title{Using Degeneracy in the Loss Landscape for Mechanistic Interpretability}

\author{%
  Lucius Bushnaq\thanks{Correspondence to Lucius Bushnaq <lucius@apolloresearch.ai>}\quad
  Jake Mendel\quad
  \AND %
  Stefan Heimersheim\quad
  Dan Braun\quad
  Nicholas Goldowsky-Dill\quad %
  \AND %
  Kaarel Hänni\thanks{Cadenza Labs}\quad
  Cindy Wu\thanks{Independent}\quad
  Marius Hobbhahn
  \AND \textmd{Apollo Research}
}

\begin{document}

\maketitle
\begin{abstract}
    
Mechanistic Interpretability aims to reverse engineer the algorithms implemented by neural networks by studying their weights and activations. An obstacle to reverse engineering neural networks is that many of the parameters inside a network are not involved in the computation being implemented by the network. These degenerate parameters may obfuscate internal structure. Singular learning theory teaches us that neural network parameterizations are biased towards being more degenerate, and parameterizations with more degeneracy are likely to generalize further. We identify 3 ways that network parameters can be degenerate: linear dependence between activations in a layer; linear dependence between gradients passed back to a layer; ReLUs which fire on the same subset of datapoints. We also present a heuristic argument that \textit{modular} networks are likely to be more degenerate, and we develop a metric for identifying modules in a network that is based on this argument.  We propose that if we can represent a neural network in a way that is invariant to reparameterizations that exploit the degeneracies, then this representation is likely to be more interpretable, and we provide some evidence that such a representation is likely to have sparser interactions. We introduce the Interaction Basis, a tractable technique to obtain a representation that is invariant to degeneracies from linear dependence of activations or Jacobians.

\end{abstract}

\newcounter{boxnumber}
\newcounter{prompt}

\setcounter{footnote}{0}

\section{Introduction}\label{sec:introduction}

Mechanistic Interpretability aims to understand the algorithms implemented by neural networks \citep{olah2017feature, elhage2021mathematical, räuker2023transparent, olah2020zoom,meng2023locating, geiger2021causal, wang2022interpretability, conmy2024towards}. A key challenge in mechanistic interpretability is that neurons tend to fire on many unrelated inputs \citep{ FUSI201666, nguyen2016multifaceted, olah2017feature, geva2021, goh2021multimodal} and any apparent circuits in the model often do not show a single clear functionality and do not have clear boundaries separating them from the rest of the network \citep{conmy2023automated, causal_scrubbing}. 

We suggest that a central problem for current methods of reverse engineering networks is that neural networks are \textit{degenerate}: there are many different choices of parameters that implement the same function \citep{wei2022deep, watanabe2009algebraic}. For example, in a transformer attention head, only the product $W_{OV} = W_O W_V$ of the $W_V$ and $W_O$ matrices influences the network's output, thus, many different choices of $W_O$ and $W_V$ are parameterizations of the same network \citep{elhage2021mathematical}. This degeneracy makes parameters and activations an obfuscated view of a network's computational features, hindering interpretability. While we have workarounds for known architecture-dependent degeneracies such as the $W_{OV}$ case, Singular Learning Theory \citep[SLT,][]{watanabe2009algebraic, watanabe2013widely} suggests that we should expect \textit{additional} degeneracy in trained networks that generalize well.

SLT quantifies the degeneracy of the loss landscape around a solution using the local learning coefficient (LLC) \citep[][]{lau2023quantifying, watanabe2009algebraic, watanabe2013widely}. More degenerate solutions lie in broader `basins' of the loss landscape, where many alternative parameterizations implement a similar function. Networks with lower LLCs are more degenerate, implement more general algorithms, and generalize better to new data \citep{watanabe2009algebraic, watanabe2013widely}. 
These predictions of SLT are only straightforwardly applicable to the global minimum in the loss landscape; a generalization is required to apply
these insights to real networks.

In this paper we make the following contributions.
First, in Section \ref{sec:slt} we propose changes to SLT to make it useful for interpretability on real networks.
Then, in Section \ref{counting} we characterize three ways in which neural networks can be degenerate.
In Section \ref{manifest}, we prove a link between some of these degeneracies and sparsity in the interactions between features.
In Section \ref{manifest/modular}, we develop a technique for searching for modularity based on its relation to degeneracy in the loss landscape.
Finally in Section \ref{lib}, we propose a practical technique for removing some of these degeneracies in the form of the interaction basis.

\section{Singular learning theory and the effective parameter count}\label{sec:slt}

If a neural network's parameterisation is degenerate, this means there are many choices of parameters that achieve the same loss. At a global minimum in the loss landscape, more degeneracy in the parametrisation implies that the network lies in a broader basin of the loss. We can quantify how broad the basin is using Singular Learning Theory [SLT, \citealt{watanabe2009algebraic, watanabe2013widely, wei2022deep}].

In Section \ref{sec:slt/background}, we provide an overview of the key concepts from SLT that we will make use of. In Section \ref{sec:slt/modifying} we explain why the tools of SLT are not completely suitable for identifying degeneracy in model internals. As a proposal to resolve some of these limitations, we introduce the \textit{behavioral loss} in Section \ref{sec:slt/behavioral},  and \textit{finite data singular learning theory} in Section \ref{sec:slt/finite_temp}. Together, these concepts will allow us to define the \textit{effective parameter count}, a measure of the number of computationally-relevant parameters in the network. If we achieved our goal of a fully parameterisation-invariant representation of a neural network, its explicit parameter count would equal its \EPC.

\subsection{Background: the local learning coefficient}\label{sec:slt/background}

The most important quantity in SLT is the \textit{learning coefficient} $\lambda$. We define a data distribution $x\sim X$ and a family of models with $\pcsymbol$ parameters, parameterised by a vector $\theta$ in a parameter space $\Theta\subseteq \mathbb R^\pcsymbol$. We also define a population loss function $L(\theta|X)$ which is normalised so that $\loss(\theta_0|X)=0$ at the global minimum $\theta_0 = \argmin_\theta L(\theta|X)$. Then $\lambda$ is defined as \citep{watanabe2009algebraic}:\footnote{See \cite{watanabe2009algebraic} for a more rigorous definition of the learning coefficient.}
\begin{align}\label{eq:learning_coeff}
    \lambda:=\lim_{\epsilon\to 0} \left[\epsilon \frac{\text{d}}{\text d \epsilon}\log \vol(\epsilon) \right]\,,
\end{align}
where 
$\vol(\epsilon)$ is the \textit{volume} of the region of parameter space $\Theta$ with loss less than $\epsilon$:
\begin{align}\label{eq:volume_function}
    \vol(\epsilon) := \int_{\{\theta \in \Theta: \,L(\theta)< \epsilon\}} \text{d}\theta 
\end{align}

The learning coefficient quantifies the way the volume of a region of low loss changes as we `zoom in' to lower and lower loss. It is a measure of basin broadness, and SLT predicts that networks are biased towards points in the loss landscape with lower learning coefficient. 

Since the loss landscape can have many different solutions with minimum loss, this definition does not necessarily single out a region corresponding to a single solution.
Therefore \cite{lau2023quantifying} introduce the \textit{local} learning coefficient (LLC, denoted by $\hat\lambda$) as a way to use the machinery of SLT to study the loss landscape geometry in the neighbourhood of a particular local minimum at $\theta^*$ by restricting the volume in the definition of the learning coefficient to a neighbourhood  of that minimum $\Theta_{\theta^*} \subset \Theta$ satisfying $\theta^* = \argmin_{\theta \in \Theta_{\theta^*}}L(\theta|X)$. Then we define the local volume:
\begin{equation}\label{eq:restricted_volume}
    \vol_{\theta^*}(\epsilon) = \int_{\{\theta \in \Theta_{\theta^*}: \,L(\theta)< L(\theta^*)+\epsilon\}} \text{d}\theta
\end{equation}
and the local learning coefficient:
\begin{equation}\label{eq:LLC}
    \hat\lambda(\theta^*) = \lim_{\epsilon\to 0} \left[\epsilon \frac{\text{d}}{\text d \epsilon}\log \vol_{\theta^*}(\epsilon) \right]\,.
\end{equation}

To see why the LLC can be thought of as counting the degeneracy in the network, consider a network with $\pcsymbol$ parameters, with $\dofsymbol$ degrees of freedom in the parameterisation (such that $\dofsymbol$ of the parameters can be freely varied together or independently, without affecting the loss). Then, we can approximate the loss by a Taylor series around the minimum:
\begin{equation}\label{eq:Taylor}
    L(\theta|X) = L(\theta^*) + \frac12 (\theta-\theta^*)^T H(\theta^*) (\theta-\theta^*) + O(||\theta-\theta^*||^3)
\end{equation}

where $H(\theta^*)$ is the Hessian at the mininum. Consider the case that all functionally relevant parameters all contribute a quadratic term to the loss to leading order, and degrees of freedom correspond to parameters which the loss does not depend on at all. In this case, \cite{Murfet2020notes} explicitly calculate the LLC, showing that it equals $\frac12(\pcsymbol-\dofsymbol)$ — i.e. the LLC counts the number of functionally relevant parameters in the model.

There is a sense that in such a model, the nominal parameter count is misleading, and if there are $\dofsymbol$ degrees of freedom then there are effectively only $\pcsymbol-\dofsymbol$ actual parameters in the model. Indeed, this is the right perspective to take for selecting a model class to fit data with. \cite{watanabe2013widely} demonstrates that for models with parameter-function maps that are not one-to-one, the Bayesian Information Criterion \citep{schwarz1978estimating}, which predicts which model fit to given data generalizes best \citep{Hoogland2023Weird}, should be modified: the parameter count of the model $\pcsymbol$ should be replaced with $2\lambda$.

In this simple example, the LLC is equal to half the rank of the Hessian at the minimum, and one might wonder if these two quantities are always related in this way. It turns out that they are only the same when the loss landscape can be written locally as a sum of quadratic terms, but this isn't always true. For example, the loss landscape could be locally quartic in some directions, or the set of points with loss equal to 0 may form complicated self intersecting shapes like a cross. In these cases, it is the LLC, not the rank of the Hessian, that measures how much freedom there is to change parameters and how much we expect a particular model to generalise.

\subsection{Modifying SLT for interpretability}\label{sec:slt/modifying}

We would like to use the local learning coefficient to quantify the number of degrees of freedom in the parameterisation of a neural network — the number of ways the parameters in a neural network can be changed while still implementing the same function, or at least a highly similar function. However, there are some obstacles to using the LLC for this purpose: 
\begin{enumerate}
    \item The LLC $\hat\lambda(\theta^*)$ measures the size of the region of equal loss around a particular local minimum $\theta^*\in\Theta$ in the loss landscape. This loss landscape is defined by a loss function and a dataset of inputs and labels. Unless the network achieves optimal loss on this dataset, points in the region could have equal loss even though they correspond to \textit{different} functions, if these functions achieve the same average performance over the dataset. We do not want our measure of the number of degrees of freedom to include different functions which achieve the same overall loss.
    \item The local learning coefficient is only well defined at a local minimum of the loss, but we frequently want to interpret neural networks that have not been trained to convergence and are not at a minimum of the loss on their training distribution. 
    \item We would like to be able to consider two very similar but not identical functions to be the same function, if they only differ in ways that can be considered noise. This is partially because, after finite training time, a network will not have fully converged on the cleanest version of an algorithm without any noise\footnote{Indeed, sometimes it is possible to remove this noise and improve performance \citep{nanda2023progress}}. However, the formal approach of SLT studies models in the limit of infinite data. This turns out to correspond to taking the limit $\epsilon\to 0$ in the definition of the LLC (equation \ref{eq:LLC}) — after infinite data, the LLC is determined by the scaling of the volume function at loss equal to $L(\theta^*)$. This means that the LLC contains information only about exact degeneracies in the parameterization — only about different parameterisations that are at the local minimum. Instead, we would prefer to work with a modified LLC which quantifies the number of parameterization choices which correspond to approximately identical functions.
\end{enumerate}

We introduce the \textit{behavioral loss} as a resolution to problems (1) and (2), and \textit{finite data SLT} as a resolution to problem (3).

\subsubsection{Behavioral loss}\label{sec:slt/behavioral}
In this section, we describe how we can define the local learning coefficient of a network to avoid problems 1 and 2 listed above. 
We want to define a new loss function and corresponding loss landscape for the sake of the SLT formalism (we do not train with this loss) such that all the parameter choices in a region with zero loss correspond to \textit{the same function} on the training dataset: the same map of inputs to outputs.
This loss function, which we call the Behavioral Loss, $L_B$, is defined with respect to an original neural network with an original set of parameters $\theta^*$, and defines how similar the function $\mathbf{f}_\theta$ implemented by a different set of parameters $\theta$ is to the original function $\mathbf f_{\theta^*}$:
\begin{equation}
    \loss_B(\theta | \theta^*, \dataset) = \frac1n \sum_{x\in\dataset}\left|\left|
        \mathbf f_\theta(x) - \mathbf f_{\theta^*}(x)
    \right|\right|^2
\end{equation}
where $\dataset$ is the training dataset and $||\mathbf v||$ denotes the $\ell^2$-norm of $\mathbf v$\footnote{We arbitrarily chose an MSE loss here, but conceptually we require a loss which is \textit{non-negative} and satisfies \textit{identity of indiscernibles}: $L=0 \iff \forall x: \mathbf f_\theta(x) = \mathbf f_{\theta^*}(x)$. For example, when studying an LLM, it may be more suitable to use KL-divergence.}.
By definition, this loss landscape always has a global minimum at the parameters the model actually uses $\theta = \theta^*$, solving problem 2 above. Additionally, parameter choices which achieve 0 behavioral loss must have the same input-output behaviour as $\mathbf f_\theta^*$ on the entire training dataset, solving problem 1.
Note that achieving zero behavioral loss relative to a model with parameters $\theta^*$ is a stricter requirement than achieving the same loss as the model with parameters $\theta^*$ on the training data. 
Therefore, the behavioral loss LLC $\hat\lambda_B$ will be equal to, or higher than the training loss LLC $\hat\lambda$.

\subsubsection{Singular learning theory at finite data}\label{sec:slt/finite_temp}
Next we want to resolve the problem that standard SLT formulae concern only the limit of infinite data when the model is certainly at a local minimum of the loss landscape. We would like to think of a neural network trained on a finite amount of data as implementing a core algorithm we are interested in reverse engineering, plus some amount of `noise' which may vary with the parameterisation and which is not important to interpret. For example, in a modular addition transformer \citep{nanda2023progress}, there are parts of the network which can be ablated to \textit{improve} loss: these parts of the network may be present because the model has not fully converged to a minimum yet. In this case, if we have two transformers trained on modular addition which have the same input-output behaviour \textit{after} we have ablated parts to improve performance, then we would like to consider these models as implemtenting the same function `up to' noise \textit{before} we ablate those parts.

In this section, we sketch how to modify SLT so that the LLC becomes a measure of how many different parameterisations implement \emph{nearly} the same function, rather than exactly the same function. In this way, we can numerically vary how much the functions two different parameterisations implement are allowed to differ from each other on the training data.

We start by explaining why SLT takes the limit $\epsilon \to 0$ in the definition of the learning coefficient (equation \ref{eq:learning_coeff}). SLT is a theory of Bayesian learning machines: learning machines which start with some prior over parameters which is nonzero everywhere $\varphi: \Theta \mapsto (0,1)$, and which learn by performing a Bayesian update on each datapoint they observe. After a dataset $\dataset_n$ of $n$ datapoints, the posterior distribution over parameters is:
\begin{align}\label{eq:posterior}    
p(\theta|\dataset_n) = \frac{e^{-n\loss(\theta|\dataset_n)}\varphi(\theta)}{p(\dataset_n)}\,.
\end{align}
where $L(\theta|\dataset_n)$ is the negative log likelihood of the dataset given the model $\mathbf f_\theta$, which we identify with the loss function when making a connection between Bayesian learning and SGD \citep{murphy2012machine}, and $p(\dataset_n)$ is a normalisation factor.

The exponential dependence on $n$ ensures that in the limit $n\to \infty$, a Bayesian learning machine's posterior is only nonzero at points of minimum loss. This means that the asymptotic behaviour of the learning machine depends only on properties of the loss landscape that are asymptotically close to having zero loss. This is the reason that we take $\epsilon\to0$ in the definition of the learning coefficient.

However, since the parameters $\theta^*$ we find after finite steps of SGD correspond to an algorithm plus noise, we want to consider the size of the region of parameter space that achieves a behavioral loss \textit{less than the noise size}. From a bayesian learning perspective, in equation \ref{eq:posterior}, we can see that for large but finite number of data points, most of the posterior concentrates around the regions of low loss, but it does not fully concentrate on the region with exactly minimum loss.

Therefore, we simply refrain from taking the limit as the loss scale $\epsilon$ goes to $0$ in the definition of the learning coefficient, and consider the learning coefficient at a particular loss scale:
\begin{align}
    \lambda(\epsilon):=\epsilon \frac{\text d}{\text d \epsilon}\log\vol(\epsilon)
\end{align}

To understand how the learning coefficient can vary with epsilon, consider an illustrative example: an extremely simple setup with a single parameter $w \in \mathbb R$, and a loss function $\loss(w) = c^2w^2 + w^4$ with $c\ll1$. This is a toy model of a scenario where there is a very small quadratic term in the learning coefficient. This term is only `visible' to the learning coefficient when we zoom in to very small loss values. To see this, we must the calculate how the volume (equation \ref{eq:volume_function}) depends on the loss scale $\epsilon$. 
For large $\epsilon\gg c^\frac14$, the quartic term dominates the loss and the region of loss less than $\epsilon$ is roughly the interval $[-\epsilon^\frac14, \epsilon^\frac14]$. 
This gives $\vol(\epsilon)\approx 2\epsilon^\frac14$ so the learning coefficient is $\lambda(\epsilon\gg c^\frac14) = \frac14$, the same as if the quadratic term were not present. 
On the other hand, for small enough $\epsilon \ll c^\frac14$, the quadratic term becomes visible: $\vol(\epsilon) \approx 2\epsilon^{\frac12}/c^2$, so $\lambda(\epsilon\ll c^\frac14) = \frac12$.

Determining how to choose an appropriate cutoff $\epsilon$ is still an open problem. We suggest that researchers choose the value of the behavioral loss cutoff in the context of the question they would like to answer. For example, if one trains multiple models with different seeds on the same task, then the appropriate loss cutoff may be on the order of the variance between the seeds.

Finally, we are able to quantify the amount of degeneracy in a neural network. We define the \textit{Effective Parameter Count} of a neural network $\mathbf f_{\theta^*}$ at noise scale $\epsilon$ as 
two times the local learning coefficient $\lambda_B(\epsilon)$ of the behavioral loss with respect to the network at noise scale epsilon. 
\begin{equation}\label{eq:peff}
\epcsymbol(\epsilon):= 2 \lambda_B(\epsilon)
\end{equation}
We conjecture that a fully parameterisation invariant representation of a neural network which captures all the behaviour up to noise scale $\epsilon$ would require $\epcsymbol(\epsilon)$ parameters.

\section{Internal structures that contribute to degeneracy}\label{counting}
In this section, we will show three ways the internal structure of neural networks can induce degrees of re-parametrization freedom $\dofsymbol$ in the loss landscape. Since $\epcsymbol=\pcsymbol-\dofsymbol$, this is equivalent to showing three ways the internal structures of neural networks determine their effective parameter count. We do not expect that these three sources of re-parametrization freedom offer a complete account of all degeneracy in real networks. They are merely a starting point for relating the degeneracy of networks to their computational structure at all.

For ease of presentation, most of the expressions in this section are only derived for the example case of fully connected networks. They can be generalized to transformers, though we do not show this explicitly here.

In Section \ref{counting/hessian}, we show a relationship between the \EPC and the \emph{dimensions} of the spaces spanned by the network's activation vectors (Section \ref{counting/activations}) and Jacobians (Section \ref{counting/jacobians}) recorded over the training data. 
In Section \ref{counting/polytopes}, we show a relationship between the number of \emph{distinct nonlinearities} implemented in a layer of the network on the training data and the \EPC.

\subsection{Activations and Jacobians}\label{counting/hessian}
In this section, we show how a network having low dimensional hidden activations or Jacobians leads to re-parametrisation freedom.

We begin by bringing the network's Hessian, which gives the first non-zero term in the Taylor expansion of the loss around an optimum (See equation \ref{eq:Taylor}) into a more convenient form. 
Each local free direction in the loss landscape corresponds to an eigenvector of the Hessian with zero eigenvalue.\footnote{The reverse does not hold, due to higher order terms in the expansion in equation \ref{eq:Taylor}. See \citep{watanabe2009algebraic, watanabe2013widely}.} Therefore, the rank of the Hessian can be used to obtain a lower bound for the learning coefficient. 

Consider the Hessian of a fully connected network, with parameters $\theta=\theta^*$, network inputs $x$ and network outputs $\mathbf f_\theta(x)$, on a behavioural loss
$\loss_B\left(\theta| \theta^*,\dataset\right)$ evaluated on a dataset consisting of $|\dataset| = n$ inputs. Using the chain rule, the Hessian at the global minimum $\theta = \theta^*$ can be written as:
\begin{equation}\label{eq:hessian}
\begin{aligned}
\left.\frac{\partial^2\loss_B\left(\theta| \theta^*,\dataset\right)}{\partial\theta^l_{i,j} \partial\theta^{l'}_{i',j'}}\right|_{\theta = \theta^*}
&=
\sum_{x\in \dataset}\sum_{k,k'}
\left.\frac{\partial^2\loss_B\left(\theta| \theta^*,\dataset\right)}{\partial{f^{\finallayer}_k}\partial{f^{\finallayer}_{k'}}}\right|_{\theta = \theta^*}
\frac{\partial f^{\finallayer}_k(x)}{\partial\theta^{l'}_{i',j'}}
\frac{\partial f^{\finallayer}_{k'}(x)}{\partial\theta^l_{i,j}}\\
&\overset{\rm MSE\ loss}{=} \frac1n \sum_{x\in \dataset}\sum_{k}
\frac{\partial f^{\finallayer}_k(x)}{\partial\theta^{l'}_{i',j'}}
\frac{\partial f^{\finallayer}_{k}(x)}{\partial\theta^l_{i,j}}
\end{aligned}
\end{equation}
for $l = 1,\dots \finallayer;\: j = 1, \dots, d^l; \: i = 1,\dots, d^{l+1}$. In the second line, we have used that the loss function is MSE from outputs at $\theta = \theta^*$ to simplify the expression, and we have also used that the first derivatives of the loss are zero at the minimum\footnote{If we were to use a different behavioural loss such as KL divergence, this would mean that the term $\left.\frac{\partial^2 L }{f^{\finallayer}_k f^{\finallayer}_{k'}}\right|_{\theta = \theta^*}$ would not be equal to $\delta_{kk'}.$ This means that different output activations (logits for a language model) would be weighted differently, but the story of this section would be largely the same.}. Thus, the Hessian is equal to a Gram matrix
of the network's weight gradients $\frac{\partial f^{\finallayer}_k}{\partial\theta^l_{i,j}}$, and linear dependence of entries of the weight gradients over the training set $\dataset$ corresponds to zero eigenvalues in the Hessian. 

We can apply the chain rule again to rewrite the gradient vector on each datapoint as an outer product of Jacobians and activations:
\begin{equation}
\begin{aligned}\label{eq:gradient_vec_decomposition}
\frac{\partial f^{\finallayer}_k(x)}{\partial\theta^l_{i,j}}
&= \frac{\partial f^{\finallayer}_k(x)}{\partial p^{l+1}_i} f^l_j(x)
\end{aligned}
\end{equation}
where the Jacobian is taken with respect to preactivations to layer $l+1$:
\begin{equation}
\mathbf{p}^{l+1}(x)=W^l \mathbf{f}^l(x).\footnote{Throughout this paper, we have added the bias into a new zeroth column of the weights and added a zeroth coordinate to the activation vector, so that $W^l_{i,0} = b^l_i$ and $\forall x: f^l_0(x) = 1$. Now $\sum_{j=0}^d W^l_{ij}f^l_j(x) = \sum_{j=1}^d W^l_{ij} + b^l_i = p^{l+1}_i$}
\end{equation}
Thus, every degree of linear dependence in the activations $f^l_j$ or Jacobians $\frac{\partial f^{\finallayer}_k}{\partial p^{l+1}_i}$ in a layer $l$ of the network also causes degrees of linear dependence in the weight gradient $\frac{\partial f^{\finallayer}_k(x)}{\partial\theta^l_{i,j}}$ of the network, potentially resulting in re-parametrisation freedom for the network. In the next two sections, we explore how linear dependence in the activations and Jacobians respectively impact the effective parameter count.
\subsubsection{Activation vectors spanning a low dimensional subspace}\label{counting/activations}
Looking at equation \ref{eq:gradient_vec_decomposition}, each degree of linear dependence of the activations $f^l_j$ in a hidden layer $l$ of width $d^l$ over the training dataset $\dataset$,
\begin{equation}\label{eq:lin_dep}
\sum_j c_j f^l_j(x)=0\,\, \forall x \in \dataset\,,
\end{equation}
corresponds to $d^{l+1}$ linearly dependent entries in the weight gradient $\frac{\partial f^{\finallayer}_k}{\partial\theta^l_{i,j}}$, $d^{l+1}$ eigenvectors of the Hessian with eigenvalue zero, and $d^{l+1}$ fully independent free directions in the loss landscape than span a fully free $d^{l+1}$ dimensional hyperplane. So the \EPC $\epcsymbol$ will be lower than the nominal number of parameters in the model $\pcsymbol$ by $d^{l+1}$ for each such degree of linear dependence in the hidden representations.

More generally, we can take a PCA of the activation vectors in layer $l$ by diagonalising the Gram matrix of activations
\begin{equation}\label{eq:G_uncentred}
\begin{aligned}
G^l&:=\frac{1}{n} \sum_{x \in \dataset} \mathbf f^l(x) {\mathbf f^l(x)}^T
\\&=:{U^l}^T D_G^l U^l
\end{aligned}
\end{equation}
If there is linear dependence between the activations on the dataset, some of the singular values (eigenvalues of $G^l$) will be zero. If we 
transform into rotated layer coordinates $\tilde{\mathbf f}^l(x)=U^l\mathbf f^l(x), \tilde W^l = W^l {U^l}^T$, then the parameters of the transformed weight matrix in rows which connect to the directions with zero variance can be changed freely without changing the product $\tilde W^l \tilde{\mathbf f}^l$.

In reality, a gram matrix of activation vectors will never have eigenvalues that are exactly 0. However, if a particular eigenvalue has size $\sqrt{\frac1n\sum_{x \in \dataset} \left({\tilde f^l}_j(x)\right)^2}=O(\epsilon^k)$ for some $\epsilon\ll 1$, the transformed parameters inside ${\tilde W^l}$ can be changed by $O(\epsilon^{\frac12 - k})$ while only impacting the loss $L$ by $O(\epsilon)$.

This suggests that, under the finite-data SLT picture introduced in Section \ref{sec:slt/finite_temp}, singular values of the set of activation vectors that are less than $\epsilon^\frac12$ for noise scale $\epsilon$ result in a lower \EPC, with $d^{l+1}$ effective parameters less for every small singular value. 
So, if we view the PCA components in a layer $l$ as the 'elementary variables' of that layer, then the fewer elementary variables the network has in total, the lower the \EPC will be.

\paragraph{Relationship to weight norm}
One might be concerned that linear dependencies between the activation vectors on the training dataset might not hold for activation vectors outside the training dataset, such that the entries of the weight matrix that we are treating as free do in fact affect the \textit{off-distribution} outputs of the network.

However, SOTA optimisers often use weight decay or $\ell^2$ weight regularisation during training to improve network generalization \citep{loshchilov2019decoupled}. This biases training towards networks with a smaller total $\ell^2$-weight norm, $||\theta||_2 = \sum_{l=1}^{\finallayer} ||W^l||_F$. 
Since the Frobenius norm $||W^l||_F$ is invariant under orthogonal transformations, the weight regularisation can equivalently be thought of as biasing training towards low $||\tilde W^l||_F$. Since the entries of $\tilde W^l$ which connect to the zero principal components do not affect the output, the training will be biased to push them to 0. 
This is an example of weight regularisation improving generalisation performance: if, at inference time, an activation vector has variation in a direction not seen during training, a regularised model ignores that component of the activation vector.

\subsubsection{Jacobians spanning a low dimensional subspace}\label{counting/jacobians}
We have shown that if the set of activation vectors in some layer have linear dependence over a dataset, then some parameters are free to vary without affecting outputs on that dataset. A similar story can be told when the Jacobians $J^l_{ij} = \frac{\partial f^{\finallayer}_i(x)}{\partial p^{l+1}_j}$ do not span the full space of the layer. As with the activations, we look for zero eigenvalues in the gram matrix of the Jacobians:
\begin{equation}\label{eq:M_pre}
\begin{aligned}
K^l &:= \frac{1}{n} \sum_{x \in \dataset}\sum_j {J^l}^T J^l
\\&=: {R^l}^T D^l_P R^l
\end{aligned}
\end{equation}
Any zero eigenvalue in this gram matrix leads to $d^{l}$ zero eigenvalues in the Hessian, analogous to the previous section. We can transform into rotated layer coordinates $\tilde W^l= R^l W^l$, $\tilde J^l= J^l R^l {R^l}^T$and the parameters of the transformed weight matrix in columns which connect to the directions with zero variance can be changed freely without changing the product $\tilde J^l \tilde W^l$.
However, unlike with the activation PCA components, the $d^{l}$ free directions in the Hessian from Jacobians spanning a low-dimensional subspace may not always correspond to $d^{l}$ full degrees of freedom in the parametrization. This is due to the potential presence of terms above second order in the perturbative expansion around the loss optimum, see equation \ref{eq:Taylor}, which can cause the loss to change if the parameters are varied along those directions despite the Hessian being zero \cite{watanabe2009algebraic}.

\paragraph{Jacobians between hidden layers} Note that we can decompose each Jacobian from layer $l$ to layer $\finallayer$ into a product of Jacobians between adjacent layers by the chain rule:
 \begin{equation}
     \frac{\partial \mathbf f^{\finallayer}(x)}{\partial \mathbf p^{l+1}_i} = 
     \frac{\partial \mathbf f^{\finallayer}(x)}{\partial \mathbf f^{\finallayer-1}}
      \frac{\partial \mathbf f^{\finallayer-1}(x)}{\partial \mathbf f^{\finallayer-2}}
      \dots
      \frac{\partial \mathbf f^{l+2}(x)}{\partial \mathbf f^{l+1}}
      \frac{\partial \mathbf f^{l+1}(x)}{\partial \mathbf p^{l+1}}\,.
\end{equation}
Thus, any rank drop in a gram matrix of Jacobians from layer $l+k$ to layer $l+k+1$ necessarily also leads to a rank drop in the gram matrix of the Jacobians from layer $l$ to layer $\finallayer$, and thus $d^{l}$ zero eigenvalues in the Hessian.

\subsection{Synchronized nonlinearities}\label{counting/polytopes}
In this section, we demonstrate a third example of internal structure that affects the \EPC of the model. 
The two examples we presented in the previous sections might be thought of as showing how the network having fewer relevant variables in its representation in a layer leads to more degeneracy. 
The example we present in this section shows how the network performing ``fewer operations'' leads to more degeneracy.

In a dense layer with piecewise linear activation functions (ReLU or LeakyReLU), the \EPC is reduced if two neurons have the same set of data points for which they are `on' and `off'. We call neurons with this property \textit{synchronized} with each other. For simplicity, in this section, we will consider a dense feedforward network with ReLU nonlinearities at each layer, and the same hidden width $d$ throughout. 

We define the neuron firing pattern 
\begin{align}
r^l_i(x) = \frac{f^l_i(x)}{p^l_i(x)}\text{\ if\ }p^l_i(x) \neq 0,\ \text{else}\ r^l_i(x) =1\,,
\end{align}
where $p^{l}_i(x)= \sum_j W^{l-1}_{i,j} f^{l-1}_j(x)$ is the preactivation of neuron $i$.
We call two neurons $i$ and $j$ synchronized if they always fire simultaneously on the training data, $r^l_i(x) = r^l_j(x)\, \forall x\in \dataset$.

\paragraph{All synchronized} As a pedagogical aid, and to demonstrate a point on how the effective parameter count is invariant to linear layer transitions, we first consider the case of all the neurons in layer $l+1$ being synchronized together in the same firing pattern $r^{l+1}(x) $. Then we can write:
\begin{equation*}
\begin{aligned}
\mathbf{f}^{l+2}(x)&=\operatorname{ReLU}\left(W^{l+1} \operatorname{ReLU}(W^l \mathbf{f}^l(x))\right)=\operatorname{ReLU}\left(W^{l+1}r^{l+1}(x) W^l \mathbf{f}^l(x)\right)\,,\\
\end{aligned}
\end{equation*}
meaning $W^l$ and $W^{l+1}$ effectively act as a single $d \times d$ dimensional matrix $\tilde{W}=W^{l+1} W^l$. Thus, any setting of the weights $W^l$ and $W^{l+1}$ that yield the same $\tilde{W}$ do not change the network's outputs on the training data, so long as we avoid changing any of the $r^{l+1}_i(x)$. 
We can ensure that the $r^{l+1}_i(x)$ do not change as we vary the weights by restricting ourselves to alternate weight matrices 
\begin{align}
W^{l+1} \to W^{l+1}C^{-1}, W^{l} \to C W^l\quad \text{with $C$ invertible and}\quad C_{i,j}\geq 0\,\forall i,j\,.
\end{align}

Note that a linear layer (without activation function, i.e. $f_i=p_i$) is just a special case of all  neurons being synchronized $\forall i,x: r^{l+1}_i(x) = 1$. 
When $W^l$ is full rank, the drop in the effective parameter count from full synchronisation is the number of parameters in layer $l$. So we see that from the perspective of the \EPC, linear transitions `do not cost anything' — including the linear transition in the model does not meaningfully increase the \EPC compared to skipping the layer entirely. We are simply passing variables to the next layer without computing anything new with them.\footnote{
See \citep{Aoyagi_2024} for a more complete treatment of effective parameter counts in deep linear networks.}

\paragraph{synchronized blocks} Now, we consider the general case of arbitrary neuron pairs in a layer being synchronized or approximately synchronized. We can organise neurons into sets $S_a, a=1,\dots a_{\text{max}}$, with the same activation patterns $r^{l+1}_{S_a}(x)$ for all neurons in the set. We call these sets synchronized \emph{blocks}. This works because synchronisation is a transitive property, if $r^{l+1}_1(x)=r^{l+1}_2(x)$ and $r^{l+1}_1(x)=r^{l+1}_3(x)$, then $r^{l+1}_1(x)=r^{l+1}_3(x)$. 

Each neuron belongs to one block, so $\sum^{a_{\text{max}}}_{a=1} \vert S_a \vert=d$. Then we have:
\begin{equation}\label{eq:commute}
\begin{aligned}
f_i^{l+2}(x)
&=\operatorname{ReLU}\left(\sum_j\sum^{a_{\text{max}}}_{a=1}r^{l+1}_{S_a}(x)\sum_{k\in S_a} W^{l+1}_{ik} W^l_{kj} f_j^l(x)\right).
\end{aligned}
\end{equation}
We can replace $W^{l+1} \to  W^{l+1} C^{-1},\: W^l \to  CW^l$, where the matrix $C$ has a block-diagonal structure
\begin{equation*}
\begin{aligned}
&C= 
\begin{pmatrix}
    C_{[1]} & & 0\\
    & \ddots & \\
    0 & & C_{[a_\text{max}]}
    
\end{pmatrix}
\quad \text{with invertible blocks}\quad C_{[a]}\in \mathbb{R}^{\vert S_a\vert \times \vert S_a\vert}\\
\text{and}\quad &C_{[a],k',k}>0 \,\forall k,k' \in (1,\dots,\vert S_a\vert) \,.
\end{aligned}
\end{equation*}

Just as we do not expect activations and gradients to have exact rank drops, we do not expect \textit{exact} neuron synchronisation to be common in real models. Instead, we can consider two neurons to be approximately synchronized if their activations only meaningfully differ on a few datapoints. Numerically, we can define:
\begin{equation}
\begin{aligned}
&\vert r^{l+1}_{a}\vert^2 :=\frac{1}{\vert \dataset \vert}\sum_{x\in \dataset}\sum_{i, i' \in S_a} \left(r^{l+1}_{i}(x) p^{l+1}_{i}(x)-r^{l+1}_{i}(x) p^{l+1}_{i'}(x)\right)^2 \;. \\
\end{aligned}
\end{equation}
If $\vert r^{l+1}_a\vert^2$ is non-zero but small, choosing different weight matrices as above will only increase the loss by an amount proportional to $O(\vert r^{l+1}_a\vert^2)$.

\paragraph{Degeneracy counting}: For each pair of synchronized neurons $r^{l+1}_i(x),r^{l+1}_{i'}(x)$, we can set a pair of off-diagonal entries $C_{k,k'}, C_{k',k}$ in $C$ to arbitrary positive values when we change the weights to $W^{l+1} \to W^{l+1} C^{-1}, W^l \to C W^l$. 
If $W^l$ is full rank, the rows $k$ and $k'$ are linearly independent, so this synchronized pair will result in two free directions in parameter space. Thus, we have as many free directions in parameter space as we have synchronized neurons. We can also count this as the number of the synchronized neurons in each block squared
\begin{align*}
    N^{l+1}=\sum^{a_{\text{max}}}_{a=1} \vert S_a \vert^2.
\end{align*}
We then see that $N^{l+1}$ is highest if all the neuron firing patterns are synchronized, and lowest when all neurons have different firing patterns.

However, $W^l$ is not always full rank. Further, if we want to combine the degrees of freedom from neuron synchronisation with other degrees of freedom from this section, we have to be careful to avoid double-counting. If the activations in layer $l$ lie in low-dimensional subspaces, then some of the $d^2$ degrees of freedom above may already have been accounted for.
If we remove those double-counted degrees of freedom and control for the rank of $W^l$, each synchronized block only provides \emph{additional} degrees of freedom equal to the dimensionality of the space spanned by the preactivations of block $S_a$ over the dataset $\dataset$ squared, which we denote
\begin{equation}\label{eq:span_def}
    s^{l+1}_a:=\text{dim}(\text{span}\{p^{l+1}_k\vert k \in S_a\})\,.
\end{equation}
So more generally, the additional amount of degeneracy the effective parameter count is lowered by will be
\begin{equation}\label{eq:sync_dof}
    N^{l+1}= \sum_a (s^{l+1}_a)^2\,.
\end{equation}

The trivial case of self-synchronisation $i=i'$ is not excluded here in this formula. 
It corresponds to the generic freedom to vary the diagonal entries of $C$, $C_{k,k}$ of a ReLU layer: scaling all the weights going into a neuron by $C_{k,k}\in \mathbb R^+$ and scaling all the weights out of the neuron by $1/C_{k,k}$ does not change network behavior.

\paragraph{Attention} A similar dynamic holds in the attention layers of transformers, with the attention patterns of different attention heads playing the role of the $\operatorname{ReLU}$ activation patterns. If two different attention heads $h_1, h_2$ in the same attention layer have synchronized attention patterns on the training data set, their value matrices $W^{h_1}_{V}, W^{h_2}_{V}$ can be changed to add elements in the span of the value vectors of one head to the other head, with the output matrices $W^{h_1}_{O}, W^{h_2}_{O}$ that project results back into the residual stream being modified to undo the change. If $W^{h_1}_{V}, W^{h_2}_{V}$ are full rank, this results in $2 d^2_{\text{head}}$ degrees of freedom in the loss landscape for each synchronized attention head, in addition to the generic $d^2_{\text{head}}$ degrees of freedom per attention head that are present in every transformer model. {If $W^{h_1}_{V}, W^{h_2}_{V}$ is not full rank, we account for this similarly as we did with the neurons above.}

\section{Interaction sparsity from parameterisation-invariance}\label{manifest}
In the introduction, we argued that if we can represent a neural network in a parameterisation-invariant way, then this representation is likely to be a good starting point for reverse-engineering the computation in the network. 
The intuition behind this claim is that in the standard representation, parts of the network which do not affect the outputs act to obfuscate and hide the relevant computational structure — once these are stripped away, computational structure is likely to become easier to see. 
One way this could manifest is through the new representation having greater \textit{interaction sparsity}. 

In this section, we demonstrate that picking the right representation can indeed lead to sparser interactions throughout the network. Specifically, we show that we can find a representation such that, for every drop in the \EPC caused by either (a) activation vectors not spanning the activation space (Section \ref{counting/activations}) or (b) neuron synchronisation (Section \ref{counting/polytopes}), there is at least one pair of basis directions in adjacent layers of the network that do not interact.

The role of this section is to provide a first example of a representation of a network which has been made invariant to some reparameterisations, and show that this representation has correspondingly fewer interactions between variables. The algorithm sketch used to find the representation here is not very suitable for selecting sparsely connected bases in practical applications, since it is somewhat cumbersome to extend to non-exact linear dependencies. We introduce a way to choose a basis for the activations spaces that is more suitable for practical applications in Section \ref{lib}.

Consider a dense feedforward network with ReLU activation functions, with $\dofsymbol$ degrees of freedom in its parameterization that arise from a combination of 
\begin{enumerate}
    \item The gram matrix of activation vectors in some layers being low rank, see Section \ref{counting/activations}.
    \item Blocks of neurons being synchronized, see Section \ref{counting/polytopes}.
\end{enumerate}
We will now show that we can find a representation of the network that
\begin{enumerate}
    \item exploits the degrees of freedom due to low-dimensional activations to sparsify interactions through a re-parametrisation. 
    \item exploits the degrees of freedom from neuron synchronisation to sparsify interactions through a coordinate transformation, without losing the sparsity gained in step 1.
\end{enumerate}

\paragraph{Sparsifying using low dimensional activations}
Here, we show how to exploit the degrees of freedom in the network due to low-dimensional activations in the input layer to sparsify interactions.

Suppose that the gram matrix of activations $\mathbf f^{(1)}(x)$ of the input layer, $G^{(1)}=\frac1n\sum_x f^{(1)}_i(x) f^{(1)}_j(x)$ is not full rank. This means that we can take a set of $\text{rank}\left(G^{(1)}\right)$ neurons as a basis for the space. This will be fewer neurons than the width $d^{(1)}$ of the input payer. Writing
\begin{equation}
\forall j \in (\text{rank}\left(G^{(1)}\right)+1, \dots, d^{(1)}): f^{(1)}_{j}=\sum_{i=1}^{\text{rank}\left(G^{(1)}\right)} (c_j)_i f^{(1)}_i\,,
\end{equation}
we can replace the weights $W^{(1)}$ with new weights 
\begin{align}
\tilde{W}^{(1)}_{ij}:=
\begin{cases}
    W^{(1)}_{ij} +\sum_{k=\text{rank}\left(G^{(1)}\right)+1}^{d^{(1)}}(c_k)_j W_{ik} & 1 \le j \le \text{rank}\left(G^{(1)}\right) \\
    0 & \text{rank}\left(G^{(1)}\right) < j \le d^{(1)} 
\end{cases} 
\end{align}
In this way we can disconnect $(d^{(1)}-\text{rank}\left(G^{(1)}\right))$ many neurons from the next layer without changing the activations in layer 2 at all on the training dataset, since $\tilde W^{(1)} \mathbf f^{(1)} = W^{(1)} \mathbf f^{(1)}$.
For every degree of linear dependence we may have had in layer $1$, we now have $d^{(2)}$ weights set to zero, where $d^{(2)}$ is the width of the second MLP layer. Since two neurons that are connected by a weight of 0 do not interact, this means that we can associate each drop in the \EPC caused by linear dependence between activations in layer 1 with a pair of nodes in the interaction graph which do not interact.
\paragraph{Sparsifying using synchronized neurons}
Now, we show that we can exploit the degrees of freedom in the network from the synchronisation of neurons in the first hidden layer to sparsify interactions without losing any of the sparsity we gained in the previous step. 

Taking the example of the second layer $\mathbf f^{(2)}$, we want to find a new coordinate basis $\hat{\mathbf f}^{(2)} = C^{(2)} \mathbf{f}^{(2)}$ in which there is at least one pair of variables $(\hat{f}_i^{(2)},f^{(1)}_j)$ that does not interact for each drop in the \EPC caused by neuron synchronisation.

To choose this basis, we start by finding all pairs of neuron firing patterns $r^l_i(x)$ in layer $2$ that are synchronized and group them into sets of synchronized blocks.
Continuing with the same notation as in Section \ref{counting/polytopes}, we denote the blocks of synchronized neurons $S_a, a\in (1,\dots,a_{\text{max}})$, with size $|S_a|$, and we use the notation $M_{[a]}$ to denote the matrix in $\mathbb R^{s_a\times s_a}$ with entries given by $M_{ij}\: \forall i,j\in S_a$. 
Then, we choose the transformation $C^{(2)}$ to be block diagonal
\begin{equation}
\begin{aligned}
&C^2= 
\begin{pmatrix}
    C^2_{[1]} & & 0\\
    & \ddots & \\
    0 & & C^2_{[a_\text{max}]}
    
\end{pmatrix}\,,
\end{aligned}
\end{equation}
with the blocks given by the inverse\footnote{Technically the pseudoinverse, because $\tilde{W}^{(1)}_{[a]}$ does not need to be invertible.} of the $|S_a|\times |S_a|$ blocks of $\tilde{W}^{(1)}$:
\begin{align}
C^{(2)}_{[a]} = \left(\tilde{W}^{(1)}_{[a]}\right)^{-1}\,,
\end{align}
\begin{equation}
\begin{aligned}
\tilde{W}^{(1)}_{[a]} := 
\begin{pmatrix}
    W^{(1)}_{\sigma_{a-1}+1,\sigma_{a-1}+1} & \cdots & W^{(1)}_{\sigma_a,\sigma_{a-1}+1}\\
    \vdots & \ddots & \vdots \\
     W^{(1)}_{\sigma_{a-1}+1,\sigma_a} & \cdots &  W^{(1)}_{\sigma_a,\sigma_a}
\end{pmatrix}
\quad \text{for}\,\sigma_a = \sum_{b=1}^a s_b\\
\end{aligned}
\end{equation}

This coordinate transformation will set one interaction to zero per drop in the \EPC caused by neuron synchronisation. To see this, we first consider that $C^{(2)}$ commutes with the nonlinearity applied between layers 1 and 2
\begin{equation}\label{eq:C_relu_commute}
    \forall x: C^{(2)} \text{ReLU}\left(W^{(1)} \mathbf f^{(1)}(x)\right) = \text{ReLU}\left(C^{(2)} W^{(1)} \mathbf f^{(1)}(x)\right)
\end{equation}
The product $\hat{W}^{(1)} = C^{(2)}\tilde W^{(1)}$ will thus have block diagonal entries equal to the identity $\hat{W}^{(1)}_{[a]} = \mathbf{I}_{\vert S_a\vert}$. This means $\hat{W}^{(1)}$ will at minimum have an additional $\sum_a {(s^{(2)}_a)}^2 - d^{(2)}$ entries that are zero — one non-interacting pair of nodes per degree of non-generic parametrization freedom caused by neuron synchronization, see equations \ref{eq:span_def}, \ref{eq:sync_dof}.
These absent interactions are distinct from those due to the activation vectors in layer 1 not spanning the full activation space we found in the previous step. Thus, the minimum absent interactions add up to be equal or greater to the degrees of freedom in the loss landscape stemming from low dimensional activations in the input layer $f^{(1)}$ or synchronized neurons in the first hidden layer $f^{(2)}$.

\paragraph{Repeat for every layer}
Now, we can repeat the previous two steps for all layers, moving recursively from the input layer to the output layer. We check if the activation vectors in layer 2 do not span the activation space and pick new weights $\tilde{W}^{(2)}$ accordingly. Then we check if any neurons in layer three are synchronized and transform $\hat{\mathbf f}^{(3)} = C^{(3)} \mathbf{f}^{(3)}$ accordingly. We repeat this for every layer in the network. 

We thus obtain new weight matrices, and a new basis for the activations of every layer in the network. Treating the new basis vectors in each layer as nodes in a graph, we can build a graph representing the interactions in the network. This graph will have two properties:
\begin{enumerate}
    \item It has at least one interaction that is zero for every drop in the \EPC introduced by neuron synchronisation or activation vectors spanning a low dimensional subspace
    \item It is invariant to reparameterisations that exploit these degeneracies. 
\end{enumerate}

\section{Modularity may contribute to degeneracy}\label{manifest/modular}

A core goal of interpretability is breaking up a neural network into smaller parts, such that we can understand the entire network by understanding the individual parts. In this section we propose a particular notion of modularity that could be used to identify these smaller parts. We argue that this notion of modularity is likely to occur in real networks due to its relation to the LLC.

The core claim of this section is that more modular networks are biased towards lower LLC. We argue that if modules in a network interact less (i.e the network is more modular) this yields a higher total degeneracy and thus a lower LLC. Each module has internal degeneracies: if two modules do not interact then the degeneracies in each are independent of each other, so the total amount of degeneracy in the network (from these modules) is at least the sum of the amount of degeneracy within each module. However, if the modules are interacting, then the degeneracies may interact with each other, and the total amount of degeneracy in the network can be less. Therefore, networks which have non- or weakly- interacting modules typically have more degeneracy and thus a lower LLC, which means that neural networks are biased towards solutions which are modular.

The argument in this section does not preclude non-modular networks from having a lower LLC than modular networks in any specific instance. Instead, this section presents an argument that, \textit{all else equal}, modularity is associated with a lower effective parameter count. This argument could fail in practice if more modularity turns out to increase the effective parameter count of models for a different reason, or if real neural networks simply do not have low-loss modular solutions.

In Section \ref{manifest/modular/types} we define interacting and non-interacting degeneracies,
and show that the total degeneracy is higher in when individual degeneracies do not interact. In Section \ref{manifest/modular/compose} we quantify how modularity affects the LLC by studying a series of increasingly realistic scenarios. First, we consider the case of twomodules which do not interact at all in Section \ref{modularity/non-interacting}. Then we explore how to modify the analysis for modules which have a small number of interacting variables in Section \ref{modularity/add-interactions}. Finally, in Section \ref{manifest/modular/log} we extend our analysis to allow for the strength of interactions to vary. We arrive at a modularity metric which can be used to search for modules in a computational graph.

\subsection{Interacting and non-interacting degeneracies}\label{manifest/modular/types}
\begin{figure}[t]
\centering
\includegraphics[width=\linewidth]{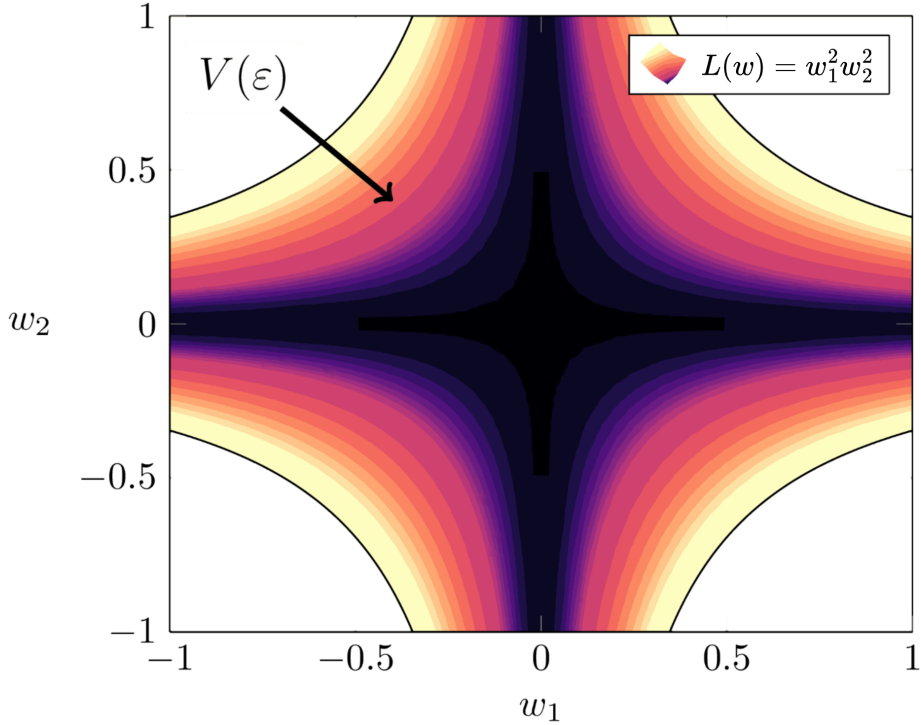}
\caption{
    Example of a loss landscape with interacting free directions, from \citep{Carroll_2023}, lightly edited.  
    The loss does not change when changing  $w_1$ alone or $w_2$ alone, so there are two free directions in the landscape. 
    However, the loss does change when changing both $w_1$ and $w_2$ together, so the set of zero loss is cross-shaped rather than spanning the whole plane. 
    Thus, despite there apparently being two free directions, the effective parameter count that characterises the dimensionality of the low loss volume is $1$ rather than $0$. 
    Non-interacting sets of parameters have no joined terms like this in the loss function, so their free directions always span full subspaces with each other.
}
\label{fig:Timaeus_cross}
\end{figure}
If a network's parameterization has a degeneracy, then there is some way the parameters of the network can change without changing the input-output behavior of the network. This change corresponds to a direction that can be traversed through the parameter space along which the behavioral loss stays zero. We call such a direction a \textit{free direction} in the parameter space. It's also possible for a parameterization to have multiple degeneracies and thus multiple free directions.

We call a set of free directions \textit{non-interacting} if traversing along one free direction does not affect whether the other directions remain free. In this case, the set of non-interacting free directions span an entire free subspace of the parameter space. In a parameter space with $\theta = (w_1,w_2,w_3)$ and loss given by $L(w_1,w_2,w_3) = w_1^2$, we are free to pick any value of $w_2$ and $w_3$ while remaining at the minimum of the loss provided that $w_1=0$. The area of constant loss is a 2-dimensional plane.

The set of free directions is called \textit{interacting} if traversing along one free direction does affect whether other directions remain free. For an extreme example, consider the loss function $L(w_1,w_2) = w_1^2 w_2^2$ (figure \ref{fig:Timaeus_cross}) at its minimum (0,0). In this case there are two free directions, but when we traverse along one free direction the other direction ceases to be free. The area of constant loss does not span a full subspace (a 2-dimensional plane); here is resembles a cross (see Figure \ref{fig:Timaeus_cross}) which is a 1-dimensional object.

We can explicitly calculate the number of degrees of freedom (the difference between the effective parameter count (equation \ref{eq:peff}) and the nominal parameter count) in each of these two loss landscapes. We find that the first landscape has two degrees of freedom but the second has only one. These are two extremes of fully interacting and fully non-interacting free directions. It is also possible to construct intermediate loss landscapes in which the number of degrees of freedom arising from two free directions is a non-integer value between 1 and 2. In general, for a given set of free directions, the lowest the effective parameter count can be is the non-interacting case.

\subsection{Degeneracies in separate modules only interact if the modules are interacting}\label{manifest/modular/compose}
In this section we quantify the increase in the effective parameter count, and equivalently the LLC,
from perfect and near-perfect modularity. We show that a network consisting of non-interacting modules
has a low effective parameter count, and that a network with modules which interact through a single variable has only a slightly higher effective paraeter count.

Consider a modular neural network $\mathbf f_\theta (x)$ consisting of two parallel modules $\model_1$ and $\model_2$. The modules take in different variables $x_1, x_2$ from the input $x=(x_1, x_2)$, and the output
of the network is the concatenation of the module outputs $\mathbf f_\theta (x) = (\model_1(x_1),\model_2(x_2))$.
We assign every activation direction in the network to either $\model_1$ or $\model_2$.

We split the parameter space $\Theta$ into 3 subspaces: $\Theta = \Theta_1 \oplus \Theta_2 \oplus \Theta_{1\leftrightarrow 2}$. $\theta_1\in\Theta_1$ are the parameters \textit{inside} $M_1$ (i.e.parameters that affect interactions between two activations within $\model_1$), $\theta_2$ is the space of the parameters inside $M_2$, and $\theta_{1\leftrightarrow 2}$ is the space of parameters which affect interactions between activations of both modules.

\subsubsection{Non-interacting case}\label{modularity/non-interacting}
We start by analyzing a network consisting of two perfectly separated modules; the values of activations in $\model_1$ have no effect on activations in $\model_2$, i.e. $\theta_{1\leftrightarrow 2}=0$ and the network
output is given by
\begin{equation}
\mathbf f_\theta (x) =(\model_1(\theta_{1},x_1),\model_2(\theta_{2},x_2)).
\end{equation}

Consider now two free directions in parameter space, where one lies entirely in $\Theta_1$, and the other lies entirely in $\Theta_2$. Since $\model_1$ and $\model_2$ share no variables and do not interact, there is no way for a change to parameters along one free direction to affect the freedom of the other direction. Therefore, \textit{one dimensional degeneracies that are in different disconnected modules must be non-interacting.}
By contrast, if $\model_1$ and $\model_2$ were connected, their free directions could interact. 

We break up the behavioral loss with respect to this network into three terms:

\begin{equation}\label{eq:behavioral_loss_factored}
    L_B(\theta|\theta^*, \dataset) = L_1(\theta_1|\theta^*_1, \dataset) + L_2(\theta_2|\theta^*_2, \dataset) + 
    L_{1\leftrightarrow 2}(\theta_1, \theta_2, \theta_{1\leftrightarrow2}|\theta^*_1, \theta^*_2, 0, \dataset)
\end{equation}
$L_1$ and $L_2$ are the parts of the behavioral loss than involve only $\theta_1$ and $\theta_2$ respectively, and $L_{1\leftrightarrow2}$ contains all the other parts. So long as we ensure $\theta_{1\leftrightarrow2}=0$, we have $L_{1\leftrightarrow2}=0$. Then a calculation shows that the overall number of degrees of freedom ($\dofsymbol = \pcsymbol - \epcsymbol$) for this behavioral loss, restricted to the subspace in which $\theta_{1\leftrightarrow2}=0$, is equal to the sum of the number of degrees of freedom in each module.

There could be additional free directions involving moving $\theta_{1\leftrightarrow 2}$ away from $0$. These free directions are not guaranteed not to interact with the free directions in each module, and our argument says nothing about how large additional contributions to the effective parameter count from varying $\theta_{1\leftrightarrow 2}$ may be.

\subsubsection{Adding in interactions between modules}\label{modularity/add-interactions}
Next, we consider the case that there are a small set of activations $\mediator_1,\dots,\mediator_m$ inside $\model_1$ that causally affect the value of some activations inside $\model_2$ (due to not all the parameters in $\theta_{1\leftrightarrow 2}$ being 0). This means that the two modules are now interacting with each other.
In that case, the only degeneracies in $\model_1$ which are guaranteed not to interact with the degeneracies in $\model_2$ are those which do not affect the value of any of the $\mediator_i$.

Picture $\model_1$ as a causal graph, where the nodes are activations and the edges are weights or nonlinearities. The nodes inside $\model_1$ are connected to the `outside' of $\model_1$ via (a) the input layer, where $\model_1$ takes in inputs, (b) the output layer, where $\model_1$ passes on its outputs, and (c) the `mediating' nodes $\mediator_i$ where variations affect what happens inside $M_2$. 
The free directions inside $\model_1$ that are guaranteed not to interact with free directions outside $\model_1$ are those directions that leave this entire \textit{interaction surface} invariant: the directions which do not change any of the mediating nodes as we traverse along them. Each mediating node that is present is an additional constraint on which free directions are guaranteed to be non-interacting. The more approximately independent nodes that are part of that interaction surface, the fewer free directions in $\model_1$ might be generically expected to satisfy these constraints.

In the previous section, we argued that the degrees of freedom of the network with noninteracting modules, restricted to the subset of parameter space in which $\theta_{1\leftrightarrow2}=0$, was equal to the sum of the degrees of freedom in each module. In this section, $\theta_{1\leftrightarrow2}^*\neq 0$, but modifying the argument to restrict to the subset of parameter space in which $\theta_{1\leftrightarrow2} = \theta_{1\leftrightarrow2}^*$ is not sufficient to fix the argument, because the degeneracies interact. 

To fix the argument, we introduce the \textit{constrained} loss function for parameters in $M_1$:
\begin{equation}\label{eq:constrained_loss}
    L_{1,C}(\theta_1|\theta_1^*, \dataset, v_1,\dots, v_m) = 
    L_1(\theta_1|\theta_1^*, \dataset) + 
    \frac1n \sum_{i=1}^m \sum_{x\in\dataset}{\left(\mediator_1(\theta_1^*, x)-\mediator_1(\theta_{1}, x)\right)}^2
\end{equation}

This loss function is the same as the part of the behavioral loss that depends only on parameters in $M_1$, except that it has extra MSE terms added to ensure that the points with very small loss also preserve the values of $\mediator_1,\dots,\mediator_m$ on all datapoints. This means its learning coefficient is higher than for the unconstrained behavioral loss. The key property of the constrained loss landscape is that free directions in are guaranteed to be non interacting with free directions in the loss landscape $L_2$. Therefore, we are able to say that the total effective parameter count of the network consisting of two interacting modules, when constrained to the subspace $\theta_{1\leftrightarrow2} = \theta_{1\leftrightarrow2}^*$, really is twice the sum of the learning coefficient for the loss function $L_2$, and for the loss function $L_{1,C}$\footnote{For simplicity in this section, we have considered the case in which nodes in $M_1$ affect nodes in $M_2$ but the converse is not true. If we wanted interactions to be bidirectional, we could modify the argument of this section by introducing a second constrained loss function $L_{2,C}$.}. 

As before, there could be additional free directions involving moving $\theta_{1\leftrightarrow2}$ away from $\theta_{1\leftrightarrow2}^*$, which may interact with the free directions in each module. Since we have not characterized the effect of these free directions on the effective parameter count, we cannot confidently conclude that networks with more separated modules reliably have lower effective parameter counts overall. For example, it may be possible that on most real-world loss landscapes, there are many more non-modular solutions than modular ones, and that typically the place in parameter space with lowest loss and lowest effective parameter count is not modular. However, we are not aware of any compelling reason why non-modular networks have some advantage in terms of having low effective parameter counts, to combat the advantage of modular networks discussed in this section.

\subsubsection{Varying the strength of an interaction}\label{manifest/modular/log}
In the precious section, we discussed the case that two modules interact via $m$ nodes. However, this model had no notion of how strong an interaction is — every node inside $\model_1$ either is not on the interaction surface, or it is, and all nodes on the interaction surface affects the nodes inside $\model_2$ the same amount. In real networks, the extent to which one activation can affect another is continuous. Therefore, we'd like to be able to answer questions like the following:

\begin{quote}
    Suppose that we have two networks both consisting of two modules, $\model_1$ and $\model_2$. In the first network, there is a single node inside $\model_1$ that strongly influences $\model_2$, and in the second there are two nodes inside $\model_1$ that both weakly influence $\model_2$. Which of these two networks is likely to have a lower effective parameter count?
\end{quote}

In this section we'll attempt to answer this question. To do so, we will make use of the notion of an effective parameter count at a finite loss cutoff $\epsilon$ (Section \ref{sec:slt/finite_temp}). We show that the magnitude of the total connections through different independent mediating nodes $\mediator_1, \mediator_2$ seems to add approximately logarithmically to determine the effective `size' of the total interaction surface between modules.

As before, we consider two modules $\model_1$ and $\model_2$, connected through a number of mediating variables $\mediator_1,\dots,\mediator_m$ that are part of $\model_1$ and which $\model_2$ depends on. Let each of these mediating variables connect to $\model_2$ through a single weight, $w_1,\dots,w_n$\footnote{We could also consider $w_i$ to be the sum of weights connecting node $\mediator_i$ to $\model_2$.}.

If $w_i$ is sufficiently small relative to the loss cutoff $\epsilon$, the connection between modules via $\mediator_i$ will be so small that it can be considered no connection at all from the perspective of interactions between free directions in different modules. This would be if the loss increases when we traverse along both free directions simultaneously by an amount that is smaller than $\epsilon$.

Quantitatively, if we traverse along a free direction in $\Theta_1$ that changes the value of $\mediator_i(\theta_1\vert x)$, then for small enough $\epsilon$ (and a network with locally smooth-enough activation functions), the resulting change in the MSE loss of the whole network $\loss$ will be proportional to $w^2_i$. If $w_i =O\left(\epsilon^{\frac{1}{2}}\right)$, that means the connection is `effectively zero' relative to the given cutoff $\epsilon$, in the sense that the volume of points with $\loss(\theta)<\epsilon$ is not substantially impacted by the terms in the loss involving $w_i$. 

Now we consider larger connections $w_i =\epsilon^{k_i}$ with $k_i \in (0,\frac{1}{2})$. 
We can model this situation by taking the size of $w_i$ into account in the constrained loss (equation \ref{eq:constrained_loss}). We define the weighted constrained loss by a sum over mean squared errors for preserving each mediating variable, weighted by the size of the variable:
\begin{align}\label{eq:weighted_constrained_loss}
    L_{1,C}(\theta_1|\theta_1^*, \theta_{1\leftrightarrow2}^*, \dataset, v_1,\dots, v_m) =  L_1(\theta_1|\theta_1^*, \dataset)
    + \frac1n \sum^m_{i=1} \epsilon^{2 k_i} \sum_{x\in\dataset}{\left(\mediator_i(\theta_1^*, x)-\mediator_i(\theta_{1}, x)\right)}^2
\end{align}
where we've made $L_{1,C}$ depend on $\theta^*_{1\leftrightarrow2}$ here because $w_i$ are parameters in $\theta^*_{1\leftrightarrow2}$. 
We are interested then in how much smaller the learning coefficient for loss landscape $L_1$ is than the learning coefficient on landscape $L_{1,C}$, as a function of loss cutoff $\epsilon$. This depends heavily on the details of the model. If the constraints are completely independent, we could perhaps model the presence of each constraint as destroying some number $\gamma_i$ of degrees of freedom compared to the model in which the constraints were not present (and the modules were fully non-interacting).
\begin{align*}
    N_{\text{eff, C}} = \epcsymbol+\sum^m_{i=1}\gamma_i\,.
\end{align*}
Now, we seek an expression for $\gamma_i$ in terms of $w_i$. Since we require $L_B(\theta)<\epsilon$, and each term in $L_B$ is positive, we also have that each constraint $MSE$ must be smaller than $\epsilon$. Rearranging, we find that
\begin{equation}\label{eq:constraints2}
\begin{aligned}
\frac1n\sum_{x\in\dataset}\left(\mediator_i(\theta_1^*,x)-\mediator_i(\theta_{1},x)\right)^2 = \epsilon^{1-2 k_i}=\tilde{\epsilon}\,.
\end{aligned}
\end{equation}
Therefore, the weights $\epsilon^{2 k_i}$ of each constraint effectively correspond to measuring the volume of points satisfying that constraint at a larger loss cutoff $\tilde{\epsilon}_i=\epsilon^{1-2 k_i}$. Now, we make an assumption that if all the weights were 1, then each constraint would be responsible for removing a similar number $\tilde \gamma$ of degrees of freedom from the network. In other words, each constraint would restrict the volume of parameter space that achieves loss less than $\epsilon$ by the same amount. Then, we can rescale this region by the factor $\epsilon^{1-2k_i}$ and we find that: 

\begin{equation}
\begin{aligned}\label{eq:log_scale}
&\gamma_i= \left(1-2 k_i\right)\tilde{\gamma}=\left(1-2\frac{\log{w_i}}{\log{\epsilon}} \right)\tilde{\gamma}\,,
\end{aligned}
\end{equation}
Therefore, the size of the logarithm of the weight $w_i$ relative to the logarithm of the cutoff $\epsilon$ becomes a prefactor reducing the number of degrees of freedom removed by constraint $i$. If $w_i=1$, then $\gamma_i=\tilde \gamma$, and if $w_i\leq\epsilon^\frac12$, then $\gamma_i=0$\footnote{For $w_i < \epsilon^\frac12$, this is effectively zero from the resolution available at loss cutoff $\epsilon$.}.

With this in mind, let us return to the question introduced at the start of this section. We will call the network with two weak interactions between modules network $A$, with two mediating nodes $\mediator_{A,1},\mediator_{A,2}$ and mediating weights $w_{A,1}=w_{A,2}$. Likewise, we denote the network with one strong interaction between modules by network $B$, with one mediating node $\mediator_{B,1}$ and one mediating weight $w_B$. How large must $w_B$ be compared to $w_{A,1}$ and $w_{A,2}$ for the interactions between modules in network $B$ to effectively remove the same number of degrees of freedom as the interactions between modules in network $A$?
Using equation \ref{eq:log_scale}, we find that
\begin{equation}\label{eq:log_add}
\begin{aligned}
    &\log{\left(\frac{w_B}{\epsilon^{\frac{1}{2}}}\right)} = \log{\left(\frac{w_{A,1}}{\epsilon^{\frac{1}{2}}}\right)}
    + \log{\left(\frac{w_{A,2}}{\epsilon^{\frac{1}{2}}}\right)}\,.
\end{aligned}
\end{equation}
So, the analysis in this section implies that connections through different mediating nodes should be considered to add together logarithmically for the purpose of estimating the number of interaction terms between degrees of freedom that live in different modules. In practice, the constraints different mediating variables impose on the loss \ref{eq:weighted_constrained_loss} are likely rarely completely independent, so this should be seen as a rough approximation to be used as a starting guess for the relevant scale of the problem.

If circuits in neural networks correspond to modules, the analysis in this section implies that we could identify circuits in networks by searching for a partition of the interaction graph of the network into modules which minimises the sum of logs of cutoff-normalised interaction strengths between modules.

\section{The Interaction Basis}\label{lib}

In this section, we propose a technique for representing a neural network as an interaction graph that is invariant to reparameterisations that exploit the freedoms in Sections \ref{counting/activations} and \ref{counting/jacobians}. The technique consists of performing a basis transformation in each layer of the network to represent the activations in a different basis that we call the \textit{Interaction Basis}.

This basis transformation removes degeneracies in actviations and Jacobians of the layer to make the basis smaller. 
The basis is also intended to `disentangle' interactions between adjacent layers as much as possible. While we do not know whether it accomplishes this in general, we do show that it does so when the layer transitions are linear. In that case, the layer transition becomes diagonal (appendix \ref{appendix/DLN}).
The interaction basis is invariant to invertible linear transformations,\footnote{Technically, as we will see, it is only invariant to up to the uniqueness of the eigenvectors of a certain matrix. But that usually just amounts to a freedom under reflections of coordinate axes in practice.} meaning the basis itself is a largely coordinate-independent object, much like an eigendecomposition (see Section \ref{lib/linear_invariance}). 

We conjecture that if we apply the interaction basis transformation to a real neural network, the resulting representation is likely to be more interpretable. In a companion paper, \experimentpaper, we develop the interaction basis further and test this hypothesis.

\subsection{Motivating the interaction basis}\label{lib/basis}
To find a transformation of network's weights and activations that is invariant to reparameterisations based on low-rank activations or low-rank Jacobians, we take equation \ref{eq:hessian}, and use equation \ref{eq:gradient_vec_decomposition} to rewrite it as

\begin{equation}
\begin{aligned}
H^{l,l'}_{ij,i'j'}(\theta^*) = \left.\frac{\partial^2\loss}{\partial\theta^l_{i,j} \partial\theta^{l'}_{i',j'}}\right|_{\theta = \theta^*}
&= \frac1n \sum_{x\in \dataset}
f^l_j(x) f^{l'}_{j'}(x)
\sum_{k}
\frac{\partial f^{\finallayer}_k(x)}{\partial p^{l+1}_i} 
\frac{\partial f^{\finallayer}_k(x)}{\partial p^{l'+1}_{i'}} .
\end{aligned}
\end{equation}

Next, we make two \textit{presumptions of independence} \citep{christiano2022formalizing}, assuming that 
\begin{enumerate}
    \item{We can take expectations over the activations and Jacobians in each layer independently}
    \item{Different layers are somewhat independent such that the Hessian eigenvectors can be largely \textit{localised} to a particular layer}
\end{enumerate}

Both of these assumptions are investigated in \cite{martens2020optimizing}, who test their validity in small networks and use it to derive a cheap approximation to the Hessian and its inverse.

This allows us to approximate the Hessian as
\begin{equation}\label{eq:Hessian_approx}
\begin{aligned}
H^{l,l'}_{ij,i'j'}(\theta^*)
&\approx
\delta_{l,l'}\left[\frac1n \sum_{x\in \dataset}
f^l_j(x) f^{l}_{j'}(x)\right]
\left[\frac1n \sum_{x\in \dataset}
\sum_{k}
\frac{\partial f^{\finallayer}_k(x)}{\partial p^{l+1}_i} 
\frac{\partial f^{\finallayer}_k(x)}{\partial p^{l+1}_{i'}} \right]\,.
\end{aligned}
\end{equation}
This effectively turns the Hessian into a product of two matrices, a gram matrix of activations in each layer 

\begin{equation}
\begin{aligned}
G^l_{jj'} = \frac1n \sum_{x\in \dataset}f^l_j(x) f^l_{j'}(x)
\end{aligned}
\end{equation}
and a Gram matrix of Jacobians with respect to the next layer's preactivations 
\begin{equation}\label{eq:K}
\begin{aligned}
K^l_{ii'} = \frac1n \sum_{x\in \dataset} \sum_{k}\frac{\partial f^{\finallayer}_k(x)}{\partial p^{l+1}_i} \frac{\partial f^{\finallayer}_k(x)}{\partial p^{l+1}_{i'}}\,.
\end{aligned}
\end{equation}
We can then find the eigenvectors of this approximated Hessian by separately diagonalising these two matrices.

We would like to find a basis for $f^l$ that excludes directions connected exclusively to zero eigenvectors of the Hessian. That is, we want to exclude directions in  $f^l$ that lie along zero eigenvectors of $G^l$, and directions that are mapped by the weight matrix $W^l$ to lie along zero eigenvectors of $K^l$.

To do this, we can backpropagate the Jacobians in equation \ref{eq:K} one step further to include the weight matrices $W^l$:
\begin{align}\label{eq:M}
    M^l_{ii'} = \frac1n \sum_{x\in \dataset} \sum_{k}\frac{\partial f^{\finallayer}_k(x)}{\partial f^l_i} \frac{\partial f^{\finallayer}_k(x)}{\partial f^l_{i'}}\,.
\end{align}
and then search for a basis in $f^l$ that diagonalises $M^l$ and $G^l$ at the same time. This basis will have one basis vector less for each zero eigenvalue of the Gram matrices of the activations and Jacobians, respectively. It will also exclude directions that lie in the null space of $W^l$.

The matrices $G^l, M^l$ are symmetric, so we can write $G^l = {U^l}^T D_G^l U^l $ and $M^l = {V^l}^T D_M^l V^l$ for diagonal $D_G, D_M$ and orthogonal $U^l, V^l$.

We can find a basis transformation $\hat{\mathbf f} = C^l \mathbf f^l$ in which both $G^l$ and $M^l$ are diagonal, in two steps: 
\begin{enumerate}
    \item Apply a \textit{whitening} transformation with respect to $G^l$: $\tilde {\mathbf f}^l = \left({D^l_G}^{1/2}\right)^+ U^l$, where the plus denotes the Moore-Penrose pseudoinverse. If the activations in layer $l$ do not span the full activation space, then the gram matrix $G^l$ must not be full rank, and some diagonal entries of $D^l_G$ are zero. By choosing this pseudoinverse, we effectively eliminate all the degeneracies from low-rank activations from our final basis. In this basis, $\tilde G^l_{ij} = \delta_{ij}$
    \item Now that $G^l$ is whitened, we can apply the transformation by $V^l$ which diagonalises $M^l$ without un-diagonalisng $G^l$ since the identity matrix is isotropic\footnote{We need to be careful which coordinate basis we are working in: the entries of $V^l$ in the basis that whitens $G^l$ and in the standard basis are different.}. At this point both $M^l$ and $G^l$ are diagonal and $C^l$ is defined up to multiplication by a diagonal matrix. We choose to multiply at the end by $\left({D^l_M}^{1/2}\right)^+$ because this eliminates degeneracies from low rank Jacobians. 
\end{enumerate}
We call the basis $\hat{\mathbf f}^l = \left({D^l_M}^{1/2}\right)^+ V^l \left({D^l_G}^{1/2}\right)^+ U^l \mathbf f^l$ the interaction basis. Basis vectors in this basis are aligned with the directions that affect the output most — in the case of a deep linear network, this means that transforming to the interaction basis provably performs an SVD of each weight matrix, resulting in basis directions which are aligned with the principal components of the output of the network (see appendix \ref{appendix/DLN}). 

We made two simplifying assumptions of independence about the Hessian to motivate this basis. While they have been used in other contexts to some success, these are still strong assumptions. Future work might investigate alternative techniques for finding a basis without these assumptions. This might only be possible with an overcomplete basis, which could connect the framework of this paper to superposition.

\subsection{Invariance to linear transformations}\label{lib/linear_invariance}
The Interaction Basis is largely a coordinate-independent object, in the sense that it is invariant under linear transformations. 
If we apply a transformation $\mathbf f^l \to 
\mathbf f_R^l = R \mathbf f^l, W^l \to W_R^l = W^l R^{-1}$ to the activation space, the final interaction basis is unchanged ($\hat{\mathbf f}^l_R = \hat{\mathbf f}^l$) for any $R \in \text{GL}_{d^l}(\mathbb R)$ up to trivial axis reflections, unless $M^l$ has repeated eigenvalues.

To show this, first note that in the whitened basis $\tilde{\mathbf f^l} = \left({D^l_G}^{1/2}\right)^+ U^l \mathbf f^l$, $G^l$ is by definition always transformed to the identity matrix
\begin{align}\tilde G^l = \left({D^l_G}^{1/2}\right)^+ G^l \left(\left({D^l_G}^{1/2}\right)^+\right)^T=\mathbf{I}\,. 
\end{align}
So if we whiten after applying the transformation $R$, $\tilde{\mathbf f}^l_R$ can only differ from $\tilde{\mathbf f}^l$ by an orthogonal transformation. Call this orthogonal matrix $Q_R$. In the whitened basis, $M_R^l$ will then be:
\begin{align}
M_R^l&=Q_R M^l Q^T_R\,.
\end{align}
So $M^l_R$ and $M^l$ only differ by an orthogonal transformation. 
The interaction basis will be the eigenbasis of $M^l_R$ and $M^l$, respectively. 
So long as a real matrix does not have degenerate eigenvalues, its eigendecomposition is basis invariant if a convention for the eigenvector normalisation is chosen, up to reflections. 
So if $M^l$ does not have multiple identical eigenvalues, the interaction basis we end up in is the same up to reflections whether we transformed with $R$ first or not. If $M^l$ does have identical eigenvalues, the basis will still be identical up to orthogonal transforms in the eigenspaces of $M^l$.

\section{Related Work}

\paragraph{Explaining generalisation} The inductive biases of deep neural networks that leads them to generalise well past their training data has been an object of extensive study \citep{zhang2021understanding}. Attempts to understand generalisation involve studying simplicity biases \citep{mingard2021sgd} and are closely related to attempts to quantify model complexity, for example via VC dimension \citep{vapnik1998statistical}, Radamacher complexity \citep{mohri2018foundations} or less widely known methods \citep{liang2019fisher, novak2018sensitivity}. This paper is heavily influenced by Singular Learning Theory \citep{watanabe2009algebraic} which uses the local learning coefficient \citep{lau2023quantifying} to quantify the effective number of parameters in the model via the flatness of minima in the loss landscape. The flatness of minima has been found to predict model generalisation, for example in \cite{li2018visualizing} for networks trained on CIFAR-10. SLT has been used to study the formation of internal structure in neural networks \citep{chen2023dynamical, hoogland2024developmental}. Understanding the internals of neural networks through the geometry of their loss landscapes was also proposed as a research direction in \citep{Hoogland_Oldenziel_Murfet_2023}.
\paragraph{Local structure of the loss landscape}
Other works have investigated the structure of neural network loss landscapes and their degeneracies around solutions found in training.
In \citep{martens2020optimizing}, it was proposed that the Hessian matrix of MLPs can be approximated as being factorisable into independent outer products of activations and gradients, and that its eigenvectors might be approximated as being localised in particular layer of the network.  
This approximation was later extended to CNNs, RNNs, and transformers in \cite{grosse2016kroneckerfactored, martens2018kroneckerfactored, grosse2023studying}. 
The approximation was used to compress models by pruning weights along directions with small Hessian eigenvalues by \cite{wang2019eigendamage}.
For deep linear networks, an analytical expression for the learning coefficient was derived in \cite{Aoyagi_2024}.
Generic degeneracies in the loss shared by all models with an MLP ReLU architecture were investigated in \cite{Carrol_2021}, and degeneracies of one hidden layer MLPs with tanh activation functions in \cite{Farrugia-Roberts2022}.
It has been found that most minima in the loss landscape can often be connected by a continuous path of minimum loss, for example in \cite{draxler2019essentially} for models trained on CIFAR.

\paragraph{Selection for modularity}
In \cite{filan2021clusterability}, it was found that MLPs and CNNs trained on vision tasks showed more modularity in the weights connecting their neurons than comparable random networks.
The observed tendency for biological networks created by evolution to be modular has been widely investigated, with various explanations for the phenomenon being proposed. \cite{Clune_2013} offer a good overview of this work for machine learning researchers, and suggests direct minimisation of connection costs between components as a primary driver of modularity in biological networks.
\cite{Kashtan2005} proposes that genetic algorithms select systems to be modular because this makes them more robust to modular changes in the systems' environments.
In \cite{liu2023seeing}, connection costs were used to regularise MLPs trained on various tasks including modular addition to be more modular in their weights, in order to make them more interpretable. 

\section{Conclusion}

We introduced the idea that the presence of degeneracy in neural networks' parameterizations
may be a source of challenges for reverse engineering them. We identified some of the sources of this degeneracy, and suggested a technique (the interaction basis) for removing this degeneracy from the representation of the network. We argued that this representation is likely to have sparser interactions, and we introduced a formula for searching for modules in the new represenation of the network based on a toy model of how modularity affects degeneracy. The follow-up paper \experimentpaper tests a variant of the interaction basis, finding that it results in representations which are sparse, modular and interpretable on toy models but it is much less useful when applied to LLMs.

\section{Contribution Statement}
LB developed the ideas in this paper with contributions from JM and KH. JM and LB developed the presentation of these ideas together. JM led the writing, with substantial support from LB, and feedback from SH and NGD. SH, DB, and NGD ran experiments to provide feedback on early versions of the interaction basis. CW ran experiments to test neuron synchronisation. 

\section{Acknowledgements}
We thank Daniel Murfet, Tom McGrath, James Fox, and Lawrence Chan for comments on the manuscript, to Dmitry Vaintrob for suggesting the concept of finite data SLT, and to Vivek Hebbar, Jesse Hoogland and Linda Linsefors for valuable discussions. Apollo Research is a fiscally sponsored project of Rethink Priorities.

\bibliography{references}
\bibliographystyle{plainnat}

\appendix
\clearpage
\section{The local interaction basis on deep linear networks}\label{appendix/DLN}

The interaction basis diagonalizes interactions between neural network layers if the layer transitions are linear. We derive this property for this \textit{local} interaction basis, a modified interaction basis in which gradients to the final layer are replaced with gradients to the immediately subsequent layer, in order to sparsify interactions between adjacent layers. In the experimental follow up to this paper, \experimentpaper discuss the local interaction basis in more detail before testing it on real networks. In this appendix, we show that the local interaction basis diagonalizes the interactions between neural network layers if the layer transitions are linear. The derivation for the non-local interaction basis follows the same structure.

In the absence of nonlinearities, a deep neural network is just a series of matrix multiplications (once an extra component is added to activation vectors with a constant value of 1, to include the bias). The sparsest way to describe this series of matrix multiplications is to multiply out the network into one multiplication, and then to rotate into the left singular basis of this matrix in the inputs, and the right singular basis in the outputs.
To see that transforming to the local interaction basis does indeed perform an SVD for deep linear networks, consider the penultimate layer of the network. We neglect mean centering to make this derivation cleaner, and start by transforming in layer $\finallayer-1$ to a basis which whitens the activations:
\begin{align*}
    f^{\finallayer} &= W^{\finallayer-1} f^{\finallayer-1}\\
    &= \underbrace{ W^{\finallayer-1} \left(U^{\finallayer-1}\right)^T \left(D^{\finallayer-1}\right)^\frac12}_{W'^{\finallayer-1}}
    \underbrace{\left((D^{\finallayer-1})^\frac12 \right)^+ U^{\finallayer-1}  f^{\finallayer-1}}_{f'^{\finallayer-1}}
\end{align*}
We've wrapped these transformations into definitions of $W'^{\finallayer-1}$ and $f'^{\finallayer-1}$.
We'll show that the other transformations perform an SVD of $W'^{\finallayer-1}$. First, we have to transform to the (uncentered) PCA basis in the final layer.
\begin{align*}
    G^{\finallayer}_{ij}&=\frac1 {n}\sum_x f^{\finallayer}_{i'}(x)f^{\finallayer}_{j'}(x)\\
    &= \frac1 {n}\sum_x W^{\finallayer-1}_{i'k}f^{\finallayer-1}_{k}W^{\finallayer-1}_{j'm}f^{\finallayer-1}_{m}\\
    G^{\finallayer} &=  W^{\finallayer-1} G^{\finallayer-1} {W^{\finallayer-1}}^T\\
    &= W'^{\finallayer-1} {W'^{\finallayer-1}}^T
\end{align*}
where we have leveraged that $G^{\finallayer-1} = {U^{\finallayer-1}}^T D^{\finallayer-1} U^{\finallayer-1}$ by definition in the last step. Writing $W'^{\finallayer-1} = U_{W'}\Sigma_{W'} V_{W'}^T$, we have that $G^L = U_{W'}\Sigma_{W'}^2 U_{W'}^T$, so $U^L = U_{W'}^T$. Since there is no layer after the final layer, the $M$ matrix is not defined for the final layer, so the LI basis in the final layer is just the PCA basis\footnote{This is also true in the nonlocal interaction basis, since $\frac{\partial f^{\finallayer}_i(x)}{\partial f^{\finallayer}_j} = \delta_{ij}$}. 
\begin{align}\label{eq:partially_rotated_W}
    \hat{f}^{\finallayer} &= U^{\finallayer} f^{\finallayer} = U_{W'}^T W'^{\finallayer-1} f'^{\finallayer-1}
\end{align}
For the final part of the transformation into the LIB, we need to calculate $M$, which depends on the jacobian from the LIB functions in the next layer to the PCA functions in the current layer:
\begin{align*}
    M^{\finallayer-1}_{j,j'} &= \frac1 {n}\sum_x \frac{\partial \hat{f}^{\finallayer}_i(x)}{\partial f'^{\finallayer-1}_j}\frac{\partial \hat{f}^{\finallayer}_i(x)}{\partial f'^{\finallayer-1}_{j'}}\\
    M^{\finallayer-1} &= {W^{\finallayer-1}}^T U_{W'} U_{W'}^T W^{\finallayer-1} = {W'^{\finallayer-1}}^T W'^{\finallayer-1}\\
    &= V_{W'}\Sigma_{W'}^2 V_{W'}^T\\
    &=: {V^{\finallayer-1}}^T \Lambda^{\finallayer-1} V^{\finallayer-1}
\end{align*}
so $V^{\finallayer-1}$ = $V_{W'}^T$ and $\Lambda^{\finallayer-1} = \Sigma_{W'}^2$. Now,
\begin{align*}
    \hat{f}^{\finallayer-1} &=  C^{\finallayer-1} f^{\finallayer-1}\\
    &= {\Lambda^{\finallayer-1}}^\frac12 V^{\finallayer-1} f'^{\finallayer-1}\\
\end{align*}
Using equation \ref{eq:partially_rotated_W}, we have:
\begin{align}
    \hat{f}^L &= U_{W'}^T W'^{\finallayer-1} V_{W'}^T \left({\Lambda^{\finallayer-1}}^\frac12\right)^+ \hat{f}^{\finallayer-1}\notag\\
    &= \Sigma_{W'}\left({\Lambda^{\finallayer-1}}^\frac12\right)^+ \hat{f}^{\finallayer-1}\\
    &= \hat{f}^{\finallayer-1}\notag
\end{align}

For layers which are not the final layer in the network, the procedure is very similar. As before, we have:
\begin{equation*}
    f'^l := \left(\left(D^l\right)^\frac12\right)^+ U^l f^l \qquad\qquad W'^l := W^l \left(U^l\right)^T \left(D^l\right)^\frac12
\end{equation*}
\begin{equation*}
    G^{l+1} = W'^l {W'^l}^T,\qquad
    U^{l+1} = U^T_{W'^l}
\end{equation*}
Now, we need to remember that $\hat{f}^{l+1} = C^{l+1}f^{l+1}$:
\begin{align*}
    f'^{l+1} 
        &= \left(\left(D^{l+1}\right)^\frac12\right)^+ U^{l+1} W'^l f'^l\\
        &= \Sigma_{W'^l}^+ U^T_{W'^l} W'^l f'^l\\
        &= V_{W'^l}^T f'^l\\
    \hat{f}^{l+1} 
        &= {\Lambda^{l+1}}^\frac12 V^{l+1} f'^{l+1}\\
        &= {\Lambda^{l+1}}^\frac12 V^{l+1} V_{W'^l}^T f'^l\\
    M^l_{j,j'} &= \frac1 {n}\sum_x \frac{\partial \hat{f}^{l+1}_i(x)}{\partial f'^l_j}\frac{\partial \hat{f}^{l+1}_i(x)}{\partial f'^l_{j'}}\\
    M^l &= V_{W'^l} {V^{l+1}}^T {\Lambda^{l+1}} V^{l+1} V_{W'^l}^T
\end{align*}
Once again, note that this expression for $M^l$ is manifestly diagonal, so
\begin{equation*}
    V^l = V^{l+1}V^T_{W'^l}, \qquad \qquad \Lambda^l = \Lambda^{l+1}
\end{equation*}
So, $V^l$ is exactly what we need in order to diagonalize the relationship, and we end up with
\begin{align}
    \hat{f}^{l+1} &= {\Lambda^{l+1}}^\frac12 V^{l+1} V_{W'^l}^T {V^l}^T {{\Lambda^l}^\frac12}^+ \hat{f}^l\notag\\
    &= \hat{f}^l
\end{align}
So, each layer of the network is the same as the final layer, which is the final activations rotated into the PCA basis, but without whitening.

\end{document}